%% file: arxiv.tex
\documentclass{article}

\usepackage[preprint]{arxiv}

\usepackage{customath}

\usepackage[utf8]{inputenc} 
\usepackage[T1]{fontenc}    
\usepackage{hyperref}       
\usepackage{url}            
\usepackage{booktabs}       
\usepackage{amsfonts}       
\usepackage{nicefrac}       
\usepackage{microtype}      
\usepackage{xcolor}         
\definecolor{mydarkblue}{rgb}{0,0.08,0.45}
\hypersetup{ %
pdftitle={},
pdfkeywords={},
pdfborder=0 0 0,
pdfpagemode=UseNone,
colorlinks=true,
linkcolor=mydarkblue,
citecolor=mydarkblue,
filecolor=mydarkblue,
urlcolor=mydarkblue,
}

\usepackage{enumitem}

\newcommand{\matt}[1]{\textcolor{black}{#1}}
\newcommand{\jj}[1]{\textcolor{black}{#1}}

\title{Learning Differentiable Surrogate Losses for Structured Prediction}

\author{%
  Junjie Yang \\
  LTCI, Télécom Paris\\
  IP Paris\\
  France \\
  \texttt{junjie.yang@telecom-paris.fr} \\
  \And
  Matthieu Labeau \\
  LTCI, Télécom Paris \\
  IP Paris \\
  France \\
  \texttt{matthieu.labeau@telecom-paris.fr} \\
  \And
  Florence d'Alché-Buc \\
  LTCI, Télécom Paris \\
  IP Paris \\
  France \\
  \texttt{florence.dalche@telecom-paris.fr} \\
}

\begin{document}

\maketitle

\begin{abstract}
Structured prediction involves learning to predict complex structures rather than simple scalar values. The main challenge arises from the non-Euclidean nature of the output space, which generally requires relaxing the problem formulation. Surrogate methods build on kernel-induced losses or more generally, loss functions admitting an Implicit Loss Embedding, and convert the original problem into a regression task followed by a decoding step. However, designing effective losses for objects with complex structures presents significant challenges and often requires domain-specific expertise. In this work, we introduce a novel framework in which a structured loss function, parameterized by neural networks, is learned directly from output training data through Contrastive Learning, prior to addressing the supervised surrogate regression problem. As a result, the differentiable loss not only enables the learning of neural networks due to the finite dimension of the surrogate space but also allows for the prediction of new structures of the output data via a decoding strategy based on gradient descent. Numerical experiments on supervised graph prediction problems show that our approach achieves similar or even better performance than methods based on a pre-defined kernel.
\end{abstract}

\input{sections/intro}
\input{sections/new-method}

\input{sections/graphs}
\input{sections/related_work}
\input{sections/experiments}

\input{sections/conclusion}

\bibliography{references}
\bibliographystyle{abbrvnat}

\appendix
\input{sections/appendix}

\end{document}

%% file: sections/intro.tex
\section{Introduction}
\label{sec:intro}
In contrast to the relatively low-dimensional output spaces associated with classification and regression tasks, Structured Prediction \citep{bakir_predicting_2007, nowozin_structured_2011} involves learning to predict complex outputs such as permutations, graphs, or word sequences. The exponential size of the structured output space, coupled with the discrete and combinatorial nature of these output objects, makes prediction challenging from both computational and statistical perspectives. Consequently, exploiting the geometry of the structured output space becomes crucial during both training and inference.

Several lines of work have been proposed in the literature to address the problem. Auxiliary function maximization methods, also known as energy-based methods \citep{lafferty_conditional_2001, tsochantaridis_large_2005, belanger_structured_2016} learn a score function that quantifies the likelihood of observing a given input-output pair. The inference is then cast as an optimization problem, where the goal is to select the \textit{most likely} output according to the learned score function. However,these approaches primarily focus on specific cases where the inference task is tractable and often struggle to go beyond structured prediction problems that can be recast as as high-dimensional multi-label classification tasks \citep{graber_deep_2018}.

On the other hand, Surrogate Regression methods, such as Output Kernel Regression \citep{weston_kernel_2003, cortes_general_2005, geurts_kernelizing_2006, kadri_generalized_2013, brouard_input_2016, el_ahmad_deep_2024}, or more generally, Implicit Loss Embedding \citep{ciliberto_consistent_2016, luise_leveraging_2019, nowak_sharp_2019, ciliberto_general_2020} define a kernel-induced loss or a loss associated to an implicit surrogate feature space. These methods solve the original structured problem by framing it as a surrogate regression problem, followed by an appropriate decoding step. By choosing suitable losses, such as squared loss induced by Graph kernels \citep{borgwardt_graph_2020} or Gromov-Wasserstein based distance \citep{vayer_optimal_2019, brogat-motte_learning_2022, yang_exploiting_2024} for graph objects, one can eventually incorporate structure-related information. 

However, designing effective loss functions for complex structured objects poses substantial challenges and often demands domain-specific expertise. Additionally, the process of crafting a tailored loss function for each distinct type of output data can be costly and inefficient. Furthermore, these loss functions are seldom differentiable, complicating the prediction task from an optimization perspective.

Contrastive Learning is an unsupervised representation learning method \matt{which operates} by \matt{minimizing the distance between the embeddings of }
positive pairs (i.e. similar pairs in the original data space) while distancing them from negative samples. Although it draws on earlier concepts, e.g. kernel learning \citep{cortes_two-stage_2010} or metric learning \citep{bellet_survey_2013}, Contrastive Learning has recently demonstrated remarkable empirical success, setting new benchmarks in fields like computer vision \citep{chechik_large_2010, chen_simple_2020, grill_bootstrap_2020}, natural language processing \citep{mnih_learning_2013, logeswaran_efficient_2018, reimers_sentence-bert_2019} or graph-based applications \citep{grover_node2vec_2016, velickovic_deep_2019, sun_infograph_2020}. As a general framework versatile across various data modalities, Contrastive Learning has been successfully applied to learning generalizable representation of \textit{input} data to benefit the downstream supervised tasks, such as classification or regression.

In this work, we apply contrastive representation learning within the Surrogate Regression framework to address a range of structured prediction problems. Rather than relying on a pre-defined loss function for structured data, we propose a new structured loss based on the squared Euclidean distance in a finite-dimensional surrogate feature space. The embedding function, parameterized by neural networks, is learned directly from \textit{output} data through Contrastive Learning. The learned embedding is utilized not only during the training phase to construct the training set for the surrogate regression task which is finite-dimensional vector-valued, but also in the decoding objective function during the inference phase, ensuring differentiability throughout. Our contributions can be summarized as follows:
\begin{itemize}
    \item We propose a new general framework for Structured Prediction, termed \frameworkfull{} (\framework{}), within which differentiable surrogate losses are learned
    through contrastive learning.
    \item Due to the \textit{explicit} nature of the output loss embedding, our framework allows the usage of neural networks to solve the surrogate regression problem, providing the expressive capacity needed for complex types of input data such as images or text.
    \item The differentiability of the learned loss function unlocks the possibility of designing a \textit{projected gradient descent} based decoding strategy to predict new structures. We particularly explore this possibility for graph objects.
    \item Through empirical evaluation, we demonstrate that our method matches or surpasses the performance of approaches using pre-defined losses on a text-to-graph prediction task.
\end{itemize}

%% file: sections/new-method.tex
\section{Structured Prediction with Explicit Loss Embedding}
In this section, we first define the problem of Structured Prediction (SP) and then briefly introduce Surrogate Regression methods with pre-defined losses. Then, we propose a novel and general framework, \framework{}, that allows learning a differentiable loss and then leveraging it at training and inference time.

\subsection{Background: Structured Prediction by Leveraging Surrogate Regression}

Let $\inputs$ be an arbitrary input space and $\outputs$ an output space of structured objects. Structured objects are composed of different components in interaction. For instance, in 
some experiments, $\bmY$ is the set of node and edge labeled graphs whose size (i.e., the number of nodes) is bounded. Given  a loss function $\Delta: \outputs \times \outputs \rightarrow \reals$, Structured Prediction is written as a classic supervised learning problem and refers to solving the following Expected Risk Minimization problem:
\begin{equation}\label{pb:general}
   \min_{f \in \bmF}  \mathbb{E}_{\rho}[\disloss(f(x),y)],
\end{equation}
by only leveraging a training dataset $\bmS_{\mathrm{train}} = \{(x_i,y_i)\}_{i=1}^n~\subset (\inputs \times \outputs)^{n}$, independently and identically drawn from a fixed but unknown probability distribution $\rho$ over $\inputs \times \outputs$ and a hypothesis space $\cali F \subseteq \set{f: \inputs \to \outputs}$.  \\
To address this problem in a general way using the same machinery, regardless of the nature of the structured output, a common approach in the SP literature is to resort to a flexible family of losses based on the squared loss, defined on a 
feature space $\surrogates$ of the structured output. Let us call $\embed: \outputs \rightarrow \surrogates$ a feature map that transforms an element of $\outputs$ into a vector in the Hilbert space $\surrogates$ endowed with the inner product $\langle\cdot, \cdot\rangle_{\surrogates}$ and the associated norm $\| \cdot \|_{\surrogates}$. We define $\disloss_{\embed}$ as the following squared distance:
\begin{align}\label{eq:ourdisloss}
     \disloss_{\embed}(y, y') \coloneqq \norm{\embed(y) - \embed(y')}{\surrogates}^2. 
\end{align}
Then it is possible to find a solution $\ferm$ to the general SP problem in Equation \ref{pb:general} by solving the surrogate vector-valued regression problem in some well-chosen hypothesis space $\bmH$:
\begin{align}\label{pb:surrogate}
   \min_{h \in \bmH} \frac{1}{n} \sum_{i=1}^n \| h(x_i) - \embed(y_i) \|^2_{\surrogates}.
\end{align}
This provides us with a surrogate estimator $\herm$. For a given input $x$, the prediction $\ferm(x)$ is then obtained by using a decoding function $d:\surrogates \rightarrow \outputs$ and a candidate set $\outputs^{c} \subset \outputs$:
\begin{equation}\label{eq:fdef}
   \ferm(x) = d \circ \herm (x) = \arg \min_{y \in \outputs^{c}} \| \herm(x) - \embed(y)\|^2_{\surrogates}.
\end{equation}
In practice, a typical approach consists in leveraging a predefined but implicit embedding, e.g., the canonical feature map of a positive definite symmetric kernel $\kernel$ over structured objects: $\embed(y):=\kernel(\cdot, y)$ in the Reproducing Kernel Hilbert Space (RKHS) defined from $\kernel$. 
It turns out that the estimator $\ferm$ obtained by this surrogate approach of the target function $f^*$, solution of Equation \ref{pb:general}, is Fisher-Consistent and its excess risk is governed by the excess risk of $\herm$ as proven in \citet{ciliberto_consistent_2016,ciliberto_general_2020}. Indeed, this family of kernel-induced losses $\Delta_{\embed}$ enjoys the Implicit Loss Embedding (ILE) property, meaning it can be defined as an inner product between features in Hilbert spaces. Various methods have been designed in the literature to exploit this property to address the SP problem. All of them make use of the kernel trick in the output space to avoid explicit computation of the features.
While these methods have achieved SOTA performance on a range of problems, including Supervised Graph Prediction (SGP), they face two main limitations. First, even though the kernel can be chosen among the large catalogue of well-studied kernels for structured data (see for instance \citet{gaertner_2008} for general structured objects and \citet{borgwardt_graph_2020} for a survey on graph kernels), this choice often depends on the prior knowledge about the problem at hand which is not always easily available. Second, the decoding problem stated as a kernel pre-image problem is limited to a search for the best solution in a set of candidates. Indeed, generally, the kernels used for structured data do not allow for differentiation.

This work aims to learn a prior-less and differentiable loss for surrogate regression through a finite-dimensional differentiable feature map $\embed: \outputs \to \reals^d$. As a result,  the novel framework also opens the door to gradient descent methods in the decoding phase.
 
\subsection{Explicit Loss Embedding}
We present here a generic framework \frameworkfull{} (\framework) where the embedding defining $\disloss_{\embed}$ is learned from data, prior to solving the regression problem. Although we highlight the use of Deep Neural Networks, the framework is flexible enough to incorporate any differentiable model, as long as the structured outputs can be expressed in a relaxed form that the differentiable model can effectively manage. We provide an instantiation of the framework in Section \ref{sec:graphs}.

\begin{figure}[t]
    \centering
    \includegraphics[width=0.9\linewidth]{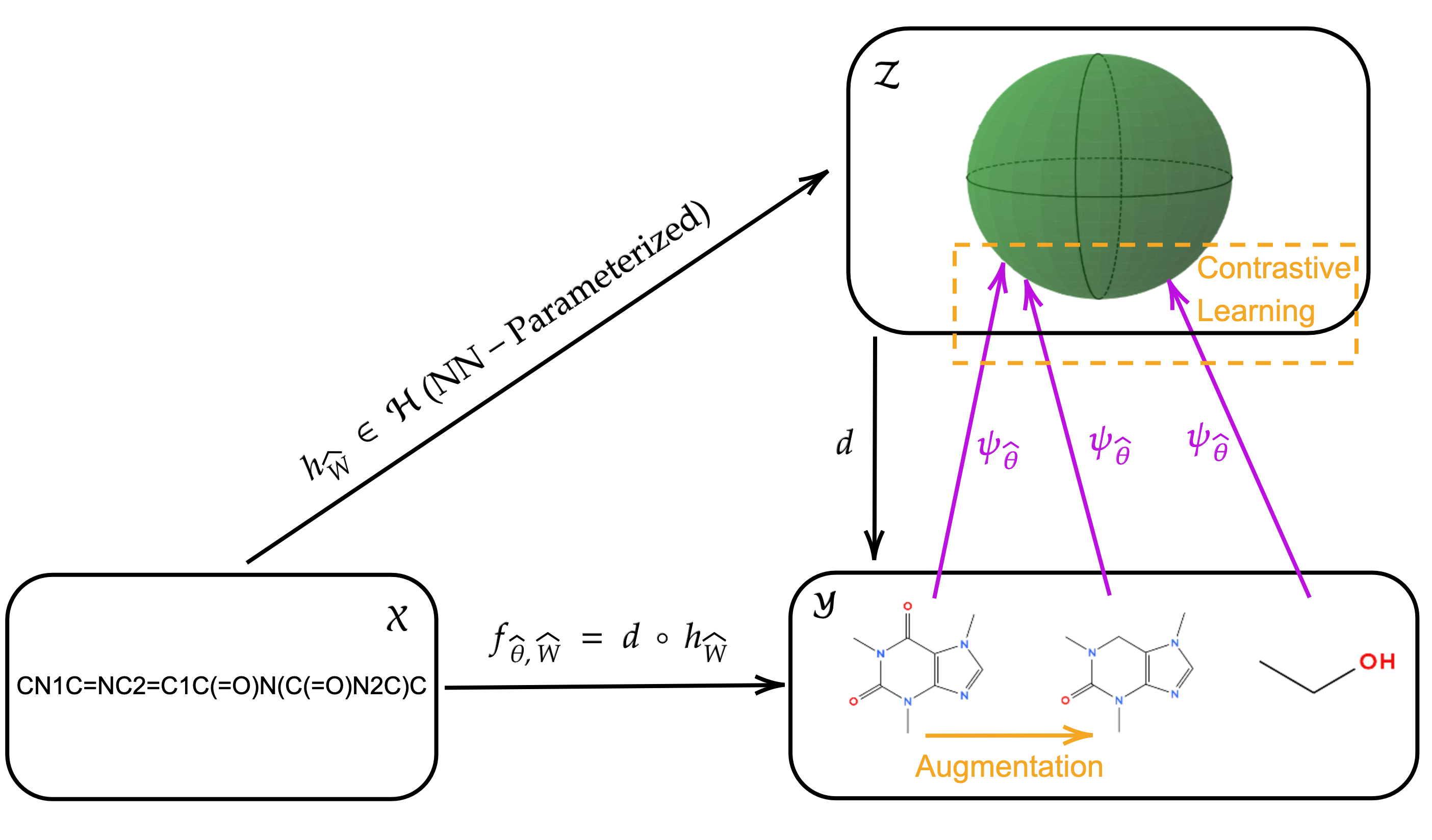}
    \caption{Illustration of \framework{} Framework.}
    \label{fig:didor}
\end{figure}

The framework \framework~illustrated in Figure \ref{fig:didor} decomposes into 3 steps:
\begin{enumerate}
\item Obtain the feature map $\hat{\embed}$ by Contrastive Learning
\item At training time, solve surrogate regression with the resulting differentiable loss $\disloss_{\hat{\embed}}(y,h(x)):= \| h(x) - \hat{\embed}(y) \|^2_{\mathbb{R}^d}$
\item At inference time, decode the prediction in the surrogate space $\hat{h}(x)$ with the differentiable loss $\disloss_{\hat{\embed}}$.
\end{enumerate}
We should note that the losses learned by ELE satisfy the ILE condition by definition. The reader can find in Section \ref{app:algo1} of the Appendix a complete view of the ELE algorithm.

\subsubsection{Feature Learning via Contrastive Learning.}

The first step involves leveraging a set of output samples $\cali{S}_{\nsamp} = \{y_i\}_{i=1}^{\nsamp} \in \outputs^{\nsamp}$ drawn from the marginal distribution $\loi{\rv Y}$ to learn the feature map $\embed$. 
We propose implementing the feature map as a parametric model such as a (deep) neural network defined by its generic parameter $\parembed$. Of paramount importance here, is the differentiability of $\embedpar$ with respect to the structured variable. Indeed, we are not only interested in $\embedpar$ as a function of $y \in \outputs$, but also require that, during inference, the feature map can be differentiable with respect to the structured variable. Therefore, we assume that we can relax a structured object $y \in \outputs$ into a continuous object $\yrelax \in \outputsrelax$ by some invertible operator $\mathcal{R}: \outputs \rightarrow \outputsrelax$. Conversely if we consider an element $\yrelax$ of $\outputsrelax$, we can convert it back into $y:=\mathcal{R}^{-1}(\yrelax)$. \jj{It should be noted that $\outputs \subset \outputsrelax$.} We give in Section \ref{sec:graphs} an example of relaxed representations of labeled graphs.

To learn $\embedpar: \outputsrelax \rightarrow \reals^d$, i.e., its parameter $\parembed$, we leverage Contrastive Representation learning (CRL) \citep{chopra_learning_2005, le-khac_contrastive_2020}, a technique that has been shown to give remarkable results in practice. In our problem, we have to define a mechanism to create similar pairs of output data $(y, y^+)$ and dissimilar pairs $(y, y^-)$ such that the CRL algorithm can learn the embedding $\embedpar$ by pulling together similar pairs $(\embedpar(y), \embedpar(y^+))$ while pushing apart dissimilar pairs $(\embedpar(y), \embedpar(y^-))$ in the feature space $\surrogates$. 

Given constructed output data $\set{(y_{\sampit}, y_{\sampit}^+, \set{y_{\sampit, k}^{-}}_{k=1}^K )}_{\sampit=1}^{\nsamp}$ where $y_{\sampit}^+$ denotes the positive sample of $y_{\sampit}$, $\set{y_{\sampit, k}^{-}}_{k=1}^K$ denotes the negative ones of $y_{\sampit}$, CRL aims at finding a NN-parameterized mapping $\psi_{\hat{\theta}}: \outputsrelax \to \sphere^{d-1} $, where $\sphere^{d-1} \coloneqq \set{x\in \reals^{d}  \;|\; \norm{x}{\reals^{d}} = 1}$ denotes the unit $(d-1)$-sphere of radius 1, such that

\begin{align}
    \hat{\theta} \in \argmin_{\theta \in \Theta} \frac{1}{\msamp}\msumsamp \ell_{\phi, \varphi} \left( \psi_{\theta}(y_i), \psi_{\theta}(y_{\sampit}^+), \set{\psi_{\theta}(y_{\sampit, k}^{-})}_{k=1}^K\right) \\
    \mathrm{with} \quad
    \ell_{\phi, \varphi} (z, z^+, \set{z_{k}^{-}}_{k=1}^K) = \phi \left(\sum_{k=1}^K \varphi\left( - \inner{z}{z^+}{\sphere^{d-1}} + \inner{z}{z_k^-}{\sphere^{d-1}}\right) \right).
    \label{eq:crl} 
\end{align}
where $\phi$ and $\varphi$ are two given monotonously increasing and differentiable scalar functions.

Since the feature learning step involves only the output space, we are no longer limited to the labeled outputs of supervised data pairs. This allows us to leverage the abundance of unsupervised output data.
\paragraph{Output Neural Network Pre-Training (ONNPT).} Given a set of additional output samples $\tilde{\cali{S}}_{\tilde{\nsamp}} = \set{y_i}_{i=1}^{\tilde{\nsamp}} \in \outputs^{\tilde{\nsamp}}$ from another distribution $\tilde{\bb P}_{\rv Y}$ close to 
$\loi{\rv Y}$, we first apply CRL on $\tilde{\cali{S}}_{\tilde{\nsamp}}$ with $T$ steps of SGD, then continue the learning procedure on the
output data  $\cali{S}_{\nsamp} = \{y_i\}_{i=1}^{\nsamp} \in \outputs^{\nsamp}$ \matt{corresponding to our labelled set}. One straightforward advantage of this  technique is that it is usually easier to get access to unsupervised output data than input-output data pairs. This pre-training 
gives us a good initialization point for the output neural network weights.
\jj{
\begin{remark}
    The paradigm of pre-training \citep{erhan_why_2010} has gained great popularity since the introduction of large language models \citep{radford_improving_2018, devlin_bert_2019, liu_roberta:_2019}. To solve a classic supervised learning problem with input-output pairs, a neural network is firstly trained on large-scale unlabelled input data with an unsupervised learning objective.  Our work extends such a strategy to the output data space in the context of structured prediction.
\end{remark}
}

Once the embedding $\embedparest$ is obtained, we are then able to define the surrogate loss $\disloss_{\embedparest}(y,h(x)):= \| \embedparest(y) - h(x) \|^2$ and inject it in Equation \ref{pb:surrogate}. 
\subsubsection{Surrogate Regression with a Learned and Differentiable Loss}
\jj{The fact that $\sphere^{d-1} \subset \reals^{d}$ is of finite dimension $d$ enables us to parameterize the hypothesis space $\cali H$ with a deep neural network $u_{\parh}: \inputs \to \reals^{d}$, whose parameter set is denoted by $\parh$, followed a $\ell_2$ normalization layer:
\begin{equation}
    \cali H \coloneqq \set{h_{\parh}(\cdot) = \frac{u_{\parh}(\cdot)}{\norm{u_{\parh}(\cdot)}{\reals^{d}}}  \mid  u_{\parh} : \inputs \to \reals^{d}, \parh \in \Parh}
\end{equation}
The choice of neural network $u_{\parh}$ depends on the nature of input data. We will refer to this network as the \textit{backbone} throughout the remainder of the paper. The hypothesis setting leads to the following optimization problem:
\begin{equation}
    \min_{\parh \in \Parh}\frac{1}{n} \sumsamp \norm{\hwpar(x_{\sampit}) - \embed_{\hat{\theta}}(y_{\sampit})}{\reals^{d}}^2.
\label{eq:surrogate_erm}
\end{equation}
Note that since for all $y \in \outputs$, $\norm{\embedparest(y)}{\reals^d} =1$, the norm boils down to the negation of the following inner product: 
$\norm{\hwpar(x)- \embed_{\hat{\theta}}(y)}{\reals^d}^2 = 2 - 2 \langle \hwpar(x), \embed_{\hat{\theta}}(y) \rangle_{\reals^d}$.}

We can then turn to a Stochastic Gradient Descent (SGD) based algorithm (e.g., Adam \citet{kingma_adam_2015}) to obtain the surrogate estimator $h_{\hatparh}$.

\subsubsection{Decoding Based Inference}
\label{sec:method_decoding}

Given an input data point $x \in \inputs$, a prediction $\hat{z}$ in the surrogate space $\surrogates$ is first obtained through $h_{\hatparh}$.
It is then decoded back to the original output space $\outputs$ through a suitable decoding function $d \in \outputs^{\surrogates}$. Our final predictor of structured objects $f_{\hat{\theta}, \hatparh}: \inputs \to \outputs$ is defined as follows:
\begin{equation}
    f_{\hat{\theta}, \hat{\parh}}(x) \coloneqq d \circ \hwhatpar(x)
\end{equation}
with
\begin{equation}
    d(\hat{z}) \in \argmin_{y \in \outputs} \norm{\embed_{\hat{\theta}}(y) - \hat{z}}{\reals^d}^2 = \argmax_{y\in \outputs} \inner{\embed_{\hat{\theta}}(y)}{\hat{z}}{\reals^d}
\label{eq:decoding}
\end{equation}
When a set of candidates $\outputs^c = \set{y_1^c, \dots, y_{n_c}^c} \subset \outputs$ of finite size $n_c$ is given, we can compute the inner product score for each candidate, and the prediction is thus given by
\begin{equation}
    f_{\hat{\theta}, \hat{\parh}}(x) = y_{\hat{j}}^c \quad \text{where} \quad  \hat{j} = \argmax_{1 \leq j \leq n_c }  \inner{\embed_{\hat{\theta}}(y_j^c)}{h_{\hat{\parh}}(x)}{\reals^d}.
\label{eq:pred_cand}
\end{equation}
\jj{We refer it as \textit{Candidate Selection} decoding strategy.
More generally, we are interested in exploiting the differentiability of the learned embedding due to the large size of the structured output space.} We hence propose to solve the decoding problem described by Equation \ref{eq:decoding} by Gradient Descent. 
\paragraph{Projected Gradient Descent Based Decoding (PGDBD).}
Instead of considering the original discrete space $\outputs$ as the constraint set of the optimization problem described by Equation \ref{eq:decoding}, we optimize the objective function over its convex hull $\outputsrelax$, and the prediction writes as a minimizer followed by the reverse operator $\mathcal{R}^{-1}$:
\begin{gather}
    f_{\hat{\theta}, \hat{\parh}}(x) = \mathcal{R}^{-1}\left(\argmin_{\yrelax\in \outputsrelax} L_x(\yrelax)\right) \quad
    \text{with} \quad L_x(\yrelax) = \norm{\embed_{\hat{\theta}}(\yrelax) - h_{\hat{\parh}}(x)}{\reals^d}^2
\end{gather}
For the constrained problem described above, the Projected Gradient Descent method (or Projected Subgradiant Descent) can be used, since  $\embed_{\hat{\theta}}$ is NN-parameterized and the squared norm is convex. With a step size $\eta$:
\begin{equation}
    \tilde{y}^{(t+1)} = \yrelax^{(t)} - \eta \nabla L_x(\yrelax^{(t)}), \quad \yrelax^{(t+1)} = \proj{\overline{\outputs}}(\tilde{y}^{(t+1)})
\end{equation}
where the projection operator $\proj{\overline{\outputs}}$ finds the point in $\overline{\outputs}$ closest to $\tilde{y}^{(t+1)}$.

The performance of a GD-based algorithm can be significantly improved using a suitable starting point. To reduce the number of gradient descent steps and obtain a better local minimum, we initialize the start point $y^{(0)}$ with the best prediction $y_{\hat{j}}^c$ of the candidate set as shown in Equation \ref{eq:pred_cand}. When the candidate set is not given, we take the output labels of the training set $\cali S_n$ as the candidate set.

Projected Gradient Descent based decoding allows us to predict new structures which are not present in the training set while reducing the burden of designing a new decoder architecture each time when a new data structure is encountered.

%% file: sections/graphs.tex
\section{Instantiation on Supervised Graph Prediction}\label{sec:graphs}
In this section, we detail how to apply our framework in a concrete, non-trivial, Structured Prediction task:  Supervised Graph Prediction (SGP). Given a data point from an arbitrary input space, SGP consists in predicting an output in a labeled graph space\matt{: we will describe this output space, how we apply the contrastive objective within it for feature learning, and how we relax it in order to apply our Projected Gradient Descent Based Decoding.}

{\textbf{Output structured space}}:
\jj{
assuming nodes can be categorized into a set of discrete labels $\nodefs$ of size $T$, edges into a set of labels $\edgefs$ of size $S$ (with no edge being considered as one of the edge labels),
each graph of at most $m_{\mathrm{max}}$ nodes can be represented by a node feature list $\mat F$ and an edge feature matrix $\mat E$, which forms a \textit{labeled graph space} $\cali G^{\nodefs, \edgefs}$ which is written as follows:
\begin{gather}
 \cali G^{\nodefs, \edgefs} \coloneqq \bigcup_{m=1}^{m_{\mathrm{max}}} \cali G_m^{\nodefs, \edgefs} \quad \text{with} \nonumber \\\cali G_m^{\nodefs, \edgefs} \coloneqq \{(\mat F, \mat E) \mid \mat F\in  \nodefs^m, \; \mat E \in \edgefs^{m\times m},  \mat E_{i, j} = \mat E_{j, i}, \forall (i, j) \in \intset{1, m}^2 \}
\label{eq:graph_space}
\end{gather}
where $F_i$ represents the label of node $i$ and $E_{i,j}$ represents the label of edge between the node $i$ and $j$.
As working on a graph space of various node sizes is difficult in practice, 
we introduce \textit{virtual} nodes $v$ into graphs so that each graph contains $m_{\mathrm{max}}$ nodes. The virtual node is treated as an additional node label, and the new graph has no edge linking to the virtual nodes. We consider thus the new padded graph space $\cali G_{m_{\mathrm{max}}}^{\nodefs \cup \set{v}, \edgefs} $ as the final output structured space $\cali \outputs$.
}

\jj{
We define a continuous relaxation $\overline{\cali G}_{m_{\mathrm{max}}}^{\nodefs \cup \set{v}, \edgefs} $ (which we will denote by $\overline{\cali G}$ for the simplification) of $\cali G_{m_{\mathrm{max}}}^{\nodefs \cup \set{v}, \edgefs}$  as follows: 
\begin{equation}
 \overline{\cali G} := \{(\mat F, \mat E) \mid \mat F\in (\simplex{T})^{m_{\mathrm{max}}}, \mat E \in (\simplex{S-1})^{m_{\mathrm{max}}\times m_{\mathrm{max}}}, \mat E_{i, j} = \mat E_{j, i}, \forall (i, j) \in \intset{1, m_{\mathrm{max}}}^2 \}
\end{equation}
where $\simplex{S-1}$ denotes a simplex of dimension $S-1$ in $\reals^{S}$. Then the relaxing operator $\mathcal{R}: \outputs \rightarrow \outputsrelax$ consists in simply mapping each node label and edge label with one-hot encoding.}

{\textbf{Contrastive learning:}}
given a graph $g = {(\mat F, \mat E)} \in \overline{\cali G}$, we adopt \textit{node dropping} (i.e., randomly discarding a certain portion of vertices along with their connections) introduced by \citet{you_graph_2020} to create the positive sample $g^+$. In practice, we replace the nodes to be dropped by the virtual nodes so that the augmented graph $g^+$ lives in the same space of $g$. The augmented graphs of random graphs in the dataset are chosen to be the negative samples $\set{g_k^-}_{k=1}^K$. Graph Neural Networks (GNNs) \citep{kipf_semi-supervised_2017, gilmer_neural_2017} are ubiquitous choices to parameterize $\psi_{\theta}: \overline{\cali G} \to \sphere^{d-1} $. We adopt Relational Graph Convolutional Networks (R-GCNs) \citep{schlichtkrull_modeling_2018}, followed by a \jj{sum pooling layer, a MLP layer and a $\ell_2$ normalization layer}. Each layer of R-GCNs does the following operation: 
\begin{equation}
    \mat F^{(l+1)} = \sigma (\sum_{s=1}^{S} \mat E_{,:,:s} \mat F^{(l)} W_{c,:,:})
\end{equation}
where $\sigma$ denotes an activation function \jj{and the tensor $W \in \reals^{S \times d \times d}$ contains the parameters of the layer except for the first layer where we have $W \in \reals^{S \times (T+1) \times d}$.} Our motivation for choosing R-GCNs is twofold: first, the above layer operation is (sub)differentiable with respect to both $\mat E$ and $\mat F$; second, it integrates the information of edge features. The output neural network $\psi_{\theta}$ are then trained to minimize the popular InfoNCE loss \citep{sohn_improved_2016, oord_representation_2018, chen_simple_2020}:
\begin{align}
    \ell_{\mathrm{InfoNCE}} (z, z^+, \set{z_{k}^{-}}_{k=1}^K) 
    & = \tau \log \left( \epsilon + \sum_{k=1}^K \exp\left(\frac{\inner{z}{z_k^-}{\sphere^{d-1}} - \inner{z}{z^+}{\sphere^{d-1}}}{\tau}\right) \right)
    \label{eq:infoNCE}
\end{align}
where $\epsilon > 0$ is an arbitrary constant and $\tau > 0$ is a scalar temperature hyperparameter.

{\textbf{Projected gradient descent based decoding:}}
The following proposition details how to conduct the projection operation $P_{\overline{\cali G}}(\cdot)$ during gradient descent.

\begin{proposition}[Projected Gradient Descent on Relaxed Graph Space]
Suppose that  $(\hat{\mat F}, \hat{\mat E})$ are the projections of $(\tilde{\mat F}, \tilde{\mat E})  \in \reals^{\mmax \times (T+1)} \times \reals^{\mmax \times \mmax \times S} $ on $\overline{\cali G}$; then, $(\hat{\mat F}, \hat{\mat E})$ are the solutions of the following optimization problems:
\begin{gather}
    \forall (i, j) \in \intset{1, \mmax}^2, \hat{\mat F}_i = \argmin_{a \in \reals^{(T+1)}} \norm{a - \tilde{\mat F}_i}{\reals^{(T+1)}}^2 \quad \mathrm{s.t.}, \; \trans{a} \vecones_{T+1} = 1, \; a \geq 0\\
     \hat{\mat E}_{i, j} = \hat{\mat E}_{j, i} = \argmin_{b \in \reals^{S}} \norm{b - (\tilde{\mat E}_{i, j} +\tilde{\mat E}_{j,i})/2}{\reals^{S}}^2 \quad \mathrm{s.t.}, \; \trans{b} \vecones_{S} = 1, \; b \geq 0 
\end{gather}
\label{prop:pgd_graph}
\end{proposition}
The proof can be found in Appendix \ref{app:proof_pgd}.
With the above proposition, we have successfully decomposed the original projection operation into a series of Euclidean projections onto the simplex, which can be independently and efficiently solved by the algorithm proposed in \citet{duchi_efficient_2008}. Furthermore, since all $\mat F$ and all $\mat E$ are of the same shape, we can compute the decoding loss in a batch fashion, and then update them through the backpropagation, which accelerates largely the decoding procedure.

\jj{The final step consists in using reverse operator $\cali R^{-1}$ to put the prediction $\overline{g} \in \overline{\cali G}$ back to the original discrete space $\cali G^{\nodefs, \edgefs}$. We first use \textit{argmax} operator to transform each feature simplex to a discrete label for both nodes and edges. Then we delete the nodes whose label  is \textit{virtual} along with their edges.}

%% file: sections/related_work.tex
\section{Related Works}
\label{sec:rw}

\subsection{Contrastive Learning of Structured Data Embedding}
While contrastive learning has long been used to facilitate learning when the output space is too large to be effectively handled, its popularity has exploded for the unsupervised learning of structured objects in the last few years. In particular, sentences (sequences of tokens) are discrete objects for which contrastive learning is a popular embedding method: an early example, Quick Thoughts \citep{logeswaran_efficient_2018} encourages sentences within the same context to be close in the embedding space while maximizing the distance between sentences from different contexts. SBERT \citep{reimers_sentence-bert_2019} selects sentences from the same section of an article as positive samples and those from different sections as negative samples. On the other hand, SimCSE \citep{gao_simcse_2021} and most of the numerous approaches that followed generate positive samples by applying a different dropout mask to intermediate representations, rather than discrete operations.
Similarly, there exists extensive literature on the unsupervised learning of graph embeddings through contrastive learning. 
Again, many variants have been explored: for creating positive samples, GraphCL \citep{you_graph_2020} designs four types of graph augmentations (node dropping, edge perturbation, attribute masking and subgraph), while MVGRL \citep{hassani_contrastive_2020} creates two views of the same graph through diffusion and sub-sampling operations.
Other approaches contrast local and global views of the graph, such as InfoGraph \citep{sun_infograph_2020}, 
which maximises mutual information between representations of 
patches of the graph.
These methods usually employ GNNs to learn such embeddings. For more details, we refer the reader to a recent survey on the contrastive learning of graphs, such as~\citet{ju_survey_2024}.

\subsection{Connections with Previous Structured Prediction Methods}
Our proposed framework 
has strong links to several previous works focusing on structured prediction. First, Input Output Kernel Regression (IOKR) \citep{brouard_input_2016} defines a scalar-valued kernel on the structured output, where the loss function corresponds to the canonical Euclidean distance induced by the kernel features. Implicit Loss Embedding (ILE) \citep{ciliberto_consistent_2016, ciliberto_general_2020} generalizes this idea to any loss function that satisfies the ILE condition. However, both methods introduce an implicit surrogate embedding space whose dimension is often unknown or can be infinite, necessitating nonparametric methods, such as those based on input operator-valued kernels \citep{micchelli_learning_2005}, to solve the surrogate regression problems. Moreover, 
IOKR and ILE are limited to tasks where the decoding problem can still be solved in a reasonable time using approximation algorithms, such as the ranking, or where a set of candidates is provided. A second set of recent works have addressed these limitations: for example, ILE-FGW \citep{brogat-motte_learning_2022} and ILE-FNGW \citep{yang_exploiting_2024} have applied the framework to highly structured graph objects, leveraging differentiable loss functions based on Optimal Transport theory. The novel graph structures can also be predicted in a gradient descent decoding way. DSOKR \citep{el_ahmad_deep_2024} reduces the dimensionality of the infinite-dimensional feature space by a sketching \citep{mahoney_randomized_2011, woodruff_sketching_2014} version of Kernel Principal Component Analysis (KPCA) \citep{scholkopf_kernel_1997}, which enables the use of neural networks to predict representations within this subspace. However, all of those methods rely on pre-defined losses, which are not systematically differentiable and computationally efficient for each data type. In contrast, our framework \framework{} allows learning a differentiable surrogate loss tailored to each data set, and the explicit nature of the output embedding permits training the input neural network without the need for additional dimensionality reduction techniques.

Finally, our Gradient Descent Based Decoding for the pre-image problem, which goes beyond the candidate set to make a prediction, is related to Structured Prediction Energy Networks (SPEN) \citep{belanger_structured_2016} if we consider $\norm{\embed_{\hat{\theta}}(\cdot) - h_{\hat{\theta'}}(x)}{\reals^d}^2$ as the energy function $E_x(\cdot)$. However, while the contribution of SPEN is limited to multi-label output spaces, we extend it to general graph output spaces. Furthermore, we also propose an initialization strategy dedicated to dealing with this more complex output space.

%% file: sections/experiments.tex
\section{Experiments}
\label{sec:exper}
\subsection{SMILES to QM9}

\matt{We demonstrate the applicability of our method through a graph prediction task, choosing to predict graphs of molecules given their string description.}

\paragraph{\textbf{Dataset}} We use the QM9 molecule dataset \citep{ruddigkeit_enumeration_2012, ramakrishnan_quantum_2014}, containing around 130,000 small organic molecules. 
These molecules were processed using RDKit\footnote{RDKit: Open-source cheminformatics. \url{https://www.rdkit.org}}, with aromatic rings converted to their Kekule form and hydrogen atoms removed. We also removed molecules containing only one atom. Each molecule contains up to 9 atoms of Carbon, Nitrogen, Oxygen, or Fluorine, along with three types of bonds: single, double, and triple. As input features, we use the Simplified Molecular Input Line-Entry System (SMILES), which are strings describing their chemical structure. We refer to the resulting dataset as \textbf{SMI2Mol}.

\paragraph{\textbf{Task}} Our task is to predict an output molecule graph $y \in \outputs $ from the corresponding input string $x \in \inputs $. 
The performance is measured using Graph Edit Distance (GED) between the predicted molecule and the true molecule, implemented by the NetworkX package \citep{hagberg_exploring_2008}. Besides showing the performance of our method \framework{} against baselines, we investigate in detail several decoding strategies to explore the usefulness of our Projected Gradient Based Decoding.

\subsubsection{Experimental Settings}

\jj{In our dataset, the space of molecules is represented as the graph space $\cali G^{\nodefs, \edgefs}$ described in Equation \ref{eq:graph_space} where $\nodefs$ is a set of atom types and $\nodefs$ is a set of bond types (including non-existence of bond). All the graphs are padded to graphs of $m_{\mathrm{max}}=9$ nodes.}

\textbf{Output embedding model}: As stated in Section \ref{sec:graphs}, we parameterize our output neural network $\psi_{\theta}$ with R-GCNs \citep{schlichtkrull_modeling_2018}.

\textbf{Output embedding learning}: 
In the default setting, contrastive learning is conducted only on the graphs from QM9 training set. In the setting of \textit{Output Neural Network Pre-training (ONNP)}, we use 
GDB-11 \citep{fink_virtual_2005, fink_virtual_2007} as the pre-training dataset, which enumerates 26,434,571 small organic molecules up to 11 atoms of Carbon, Nitrogen, Oxygen, or Fluorine. Hence, we pad all the molecules, including those from QM9, into the graphs of size $m_{\mathrm{max}}=11$.

\textbf{Input space and regression model}: To parameterize the output neural network $h_{\theta'}$, we use a multi-layer Transformer encoder \citep{vaswani_attention_2017}. The SMILES strings are tokenized into character sequences as inputs for the Transformer encoder, with the maximum length set to 25. 

\textbf{Decoding}:  
If unspecified, the decoding is made by looking for the most appropriate candidate in $\outputs^c$, as described at the beginning of Section~\ref{sec:method_decoding}.
To experiment with our proposed \textit{Projected Gradient Descent Based Decoding (PGDBD)}, we look into three decoding configurations:
\setlist{nolistsep}
\begin{enumerate}[noitemsep,leftmargin=*]
\item Candidate selection from $\outputs^c$, as above.
\item The predicted molecule is obtained using PGDBD, initialized with a random molecule from $\outputs^c$.
\item The predicted molecule is obtained using PGDBD, initialized with the best candidate molecule obtained from strategy 1.
\end{enumerate}

We create five dataset splits using different seeds from the full set of SMILES-molecule pairs. Each split contains 131,382 training samples, 500 validation samples, and 2,000 test samples. The details about the values of the hyperparameters or their search ranges can be found in Appendix \ref{app:hyper_para}.

\subsubsection{Results}

\begin{table}[t!]
\caption{Graph Edit Distance of different methods on SMI2Mol test set.}
\begin{center}
\begin{tabular}{lcc}
\toprule
 & GED w/o edge feature $\downarrow$ &  GED w/ edge feature $\downarrow$ \\
\midrule
SISOKR & $3.330 \pm 0.080$ & $4.192 \pm 0.109$ \\
NNBary-FGW & $5.115 \pm 0.129$ & - \\
Sketched ILE-FGW & $2.998 \pm 0.253$  & - \\
DSOKR & $\boldsymbol{1.951 \pm 0.074}$ & $2.960 \pm 0.079$\\
\midrule
\framework{}  & $2.305 \pm 0.033$ & $2.444 \pm 0.039$ \\
\framework{} w/ ONNPT  & $2.164 \pm 0.058$ & $2.291 \pm 0.083$\\
\framework{} w/ ONNPT + PGDBD & $2.131 \pm 0.075 $ & $\boldsymbol{2.252 \pm 0.102}$\\
\bottomrule
\end{tabular}
\end{center}
\label{tab:expr_s2m_5s}
\end{table}

\paragraph{\textbf{Comparison with Baseline Models.}}
Our method is benchmarked against SISOKR \citep{el_ahmad_sketch_2024}, NNBary-FGW \citep{brogat-motte_learning_2022}, ILE-FGW \citep{brogat-motte_learning_2022},
and DSOKR \citep{el_ahmad_deep_2024}. The results are presented in Table \ref{tab:expr_s2m_5s}. From them, we observe that \framework{} obtains better performance in terms of GED with edge features and competitive performance in terms of GED without edge features. The results also demonstrate the relevance of pre-training the output neural network with additional output data. However, there is no obvious performance improvement provided by PGDBD; we hence perform supplementary experiments with supplementary decoding strategies.

\begin{figure}[ht]
    \centering
    \begin{subfigure}{.45\textwidth}
        \centering
        \includegraphics[width=\linewidth]{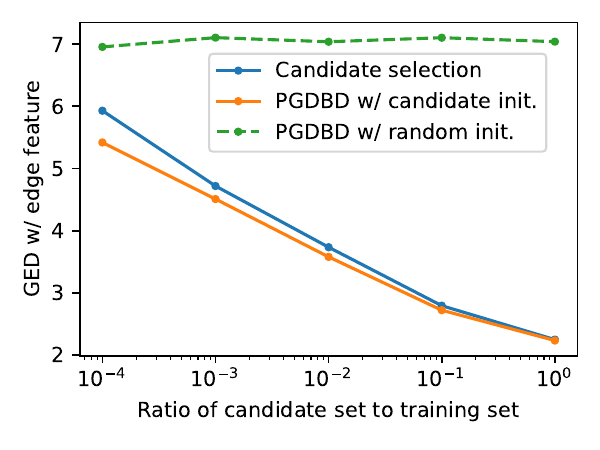}
    \end{subfigure}
    \begin{subfigure}[b]{.45\textwidth}
        \centering
        \includegraphics[width=\linewidth]{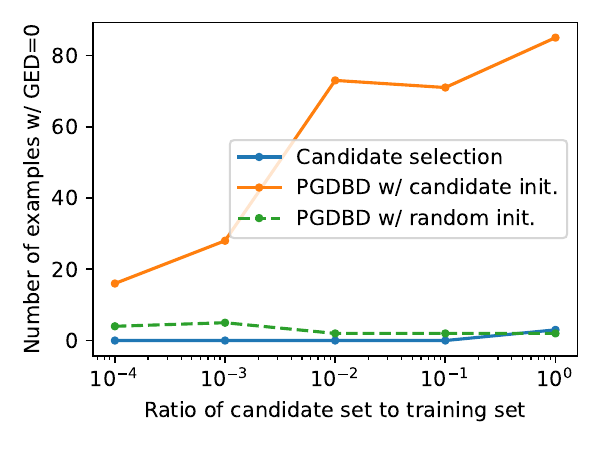}
    \end{subfigure}
    \caption{(Left) The GED with edge features of DIDOR under different decoding strategies with various sizes of candidate set. (Right) The number of predictions whose GED with the ground truth is zero, obtained by DIDOR under different decoding strategies with various sizes of candidate set.}
    \label{fig:pgdbd}
\end{figure}

\paragraph{\textbf{Study of Projected Gradient Descent Based Decoding Strategy.}} In this section, we study our proposed PGDBD strategy under the three configurations outlined in the previous section.
At the same time, we vary the size of the candidate set: in each experiment, a specific proportion of the training molecules is used to form the candidate set, controlled by a predefined ratio. In addition to the Graph Edit Distance (GED) metric, we also calculate the number of perfectly predicted outputs, which are examples for which the GED between the predicted molecule and the ground truth molecule is zero. Results on the test set are shown in Figure \ref{fig:pgdbd}.

From the GED values displayed in the left figure, we can make two key observations.
First, the performance of the PGDBD strategy improves significantly when a good graph is selected from the candidate set for initialization, rather than being drawn randomly. Second, when the size of the candidate set is limited, PGDBD enhances the quality of predictions by refining the selected candidate molecules. However, this effect becomes less pronounced as the candidate set size increases. We conjecture that a candidate graph with the highest inner product score may already be situated in a favorable local minimum, especially when considering the entire training set as $\outputs^c$.
We also posit that the highly non-convex nature of $\psi_{\theta}$ renders the optimization difficult.
The right figure, on the other hand, demonstrates the ability of our proposed PGDBD strategy to predict novel graph structures, which is impossible by simply selecting the best candidate.
\begin{figure}[t]
    \centering
    \includegraphics[width=\linewidth]{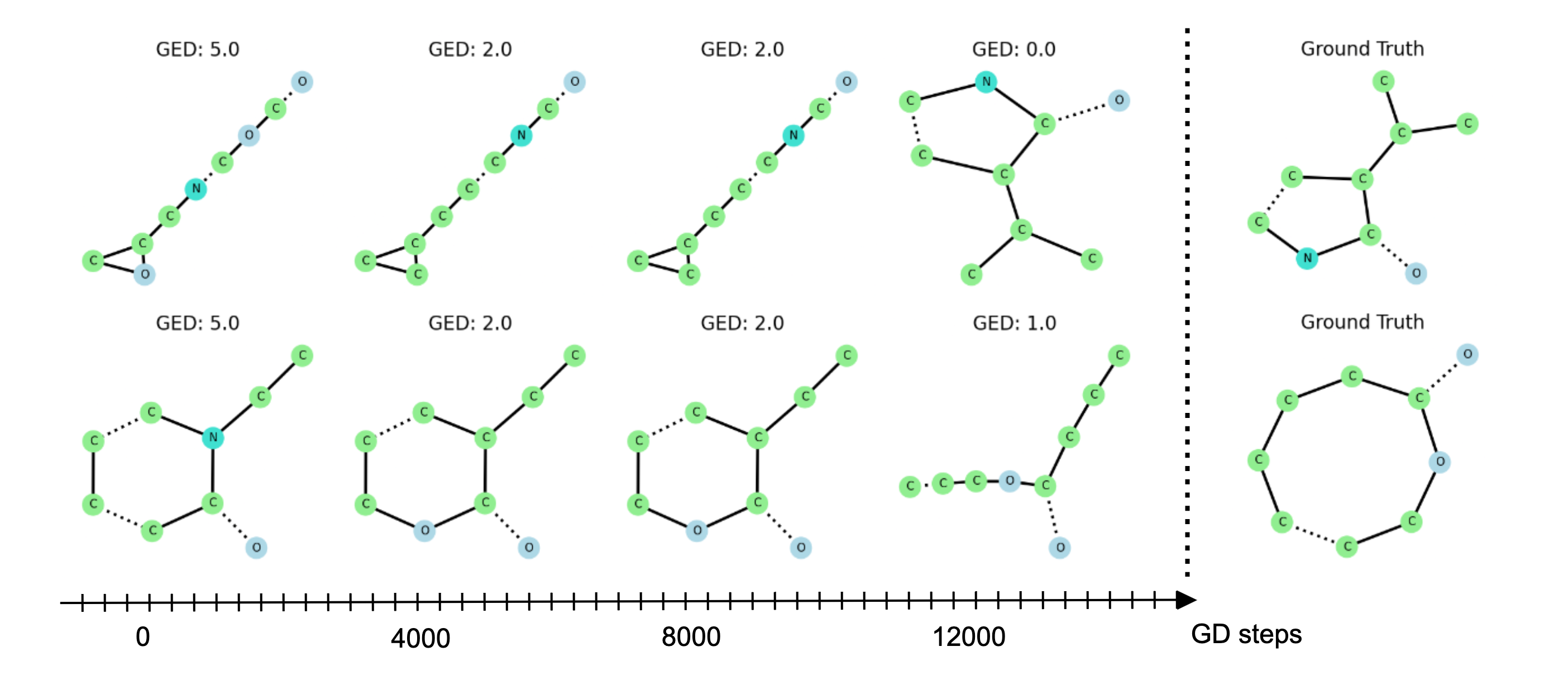}
    \caption{Update of the predicted molecules along with the projected gradient descent. }
    \label{fig:pgdbd_update}
\end{figure}
Finally, we display in Figure \ref{fig:pgdbd_update} some examples of how predicted molecules, as parameters, initialized with a random molecule, are updated through the projected gradient descent steps. We can clearly observe a phenomenon of convergence even in its original discrete space.

%% file: sections/conclusion.tex
\section{Conclusion}
\label{sec:conclusion}
In this work, we have introduced a novel framework, \frameworkfull{}, for addressing Structured Prediction tasks by leveraging contrastive learning within the Surrogate Regression setting. By building a differentiable surrogate loss from explicit embeddings learned directly from output data, our method circumvents the need for manually defining complex and often non-differentiable loss functions tailored to each structured outputs. Our empirical results, in the context of text-to-graph predictions, demonstrate that \framework{} performs on par with or surpasses other surrogate regression methods reliant on pre-defined losses. Furthermore, taking advantage of the differentiability of the surrogate loss, we propose a projected gradient descent based technique for decoding. Despite the challenge of solving a highly non-convex constrained optimization problem, our technique demonstrates the preliminary ability to predict novel graph structures, opening the door for enhancing prediction quality by incorporating more advanced optimization techniques in future work.

%% file: sections/appendix.tex
\section{Proof of Proposition \ref{prop:pgd_graph}}
\label{app:proof_pgd}
\begin{proof}
The projection on such graph space of a point $\tilde{y} = (\tilde{\mat E}, \tilde{\mat F})$ solves actually the follwoing optimization problem:
\begin{align}
    (\hat{\mat F}, \hat{\mat E}) &= \proj{\overline{\cali G}} ((\tilde{\mat E}, \tilde{\mat F})) \\
    & = \argmin_{\mat F\in (\simplex{T})^{m_{\mathrm{max}}}, \mat E \in (\simplex{S-1})^{m_{\mathrm{max}}\times m_{\mathrm{max}}}, \mat E_{i, j} = \mat E_{j, i}} \norm{\mat E - \tilde{\mat E}}{F}^2 + \norm{\mat F - \tilde{\mat F}}{F}^2 \\
    & = \argmin_{\mat F\in (\simplex{T})^{m_{\mathrm{max}}}, \mat E \in (\simplex{S-1})^{m_{\mathrm{max}}\times m_{\mathrm{max}}}, \mat E_{i, j} = \mat E_{j, i}} \sum_{i, j} \norm{\mat E_{i, j} - \tilde{\mat E}_{i, j}}{\reals^S}^2 + \sum_{i} \norm{\mat F_i - \tilde{\mat F}_i}{\reals^{T+1}}^2
\end{align}
We can easily conclude that for any $(i, j) \in \intset{1, \mmax}^2$, we have 
\begin{gather}
        \hat{\mat F}_i = \argmin_{a \in \reals^{T+1}} \norm{a - \tilde{\mat F}_i}{\reals^{T+1}}^2 \quad \mathrm{s.t.}, \; \trans{a} \vecones_{T+1} = 1, \; a \geq 0\\
        \hat{\mat E}_{i, j} = \hat{\mat E}_{j, i} = \argmin_{b \in \reals^S} \norm{b - \tilde{\mat E}_{i, j}}{\reals^S}^2 + \norm{b - \tilde{\mat E}_{j, i}}{\reals^S}^2 \quad \mathrm{s.t.}, \; \trans{b} \vecones_{S} = 1, \; b \geq 0 \label{eq:e_proj_simplex}
\end{gather}
Furthermore, it can be shown from Equation \ref{eq:e_proj_simplex} that 
\begin{equation}
    \hat{\mat E}_{i, j} = \hat{\mat E}_{j, i} = \argmin_{b \in \reals^S} \norm{b - (\tilde{\mat E}_{i, j} +\tilde{\mat E}_{j,i})/2}{\reals^S}^2 \quad \mathrm{s.t.}, \; \trans{b} \vecones_{S} = 1, \; b \geq 0 
\end{equation}
which projects the point $(\tilde{\mat E}_{i, j} +\tilde{\mat E}_{j,i})/2$ on the simplex space.
\end{proof}

\section{Pseudo-code of the ELE algorithm}
\label{app:algo1}

\begin{algorithm}[!t]
    \caption{\frameworkfull{} (\framework{})}
    \label{alg:didor}
    \SetKwInOut{Input}{input}
    \SetKwInOut{Init}{init}
    \SetKwInOut{Output}{output}
    \Input{Training set$\{(x_i, y_i)\}_{i=1}^n$, validation set$\{(x_i^{\mathrm{val}}, y_i^{\mathrm{val}})\}_{i=1}^{n_{\mathrm{val}}}$, test inputs $\{x_i^{\mathrm{te}}\}_{i=1}^{n_{\mathrm{te}}}$, candidate set $\{y_i^{\mathrm{c}}\}_{i=1}^{n_{\mathrm{c}}}$, additional output training set $\set{y_i}_{i=1}^{\tilde{\nsamp}}$}
    \Output{Predictions $\set{\hat{y}_i}_{i=1}^{n_{\mathrm{te}}}$ of the test inputs}
    \Init{Number of pre-training steps of the output NN $T_1$, number of PGDBD steps $T_2$}
    \vspace{0.2cm}
    \tcp{1. Training}
    \tcp{1.a. Training of output NN $\embed_{\parembed}$: contrastive learning of the explicit embedding}
    $\bullet$ Pre-Train output neural network $\embed_{\parembed}$ on $\set{y_{\sampit}}_{\sampit=1}^{\tilde{\nsamp}}$ with $T_1$ setps of gradient descent.

    $\bullet$ Fine-Tune output neural network $\psi_{\theta}$ on $\set{y_{\sampit}}_{\sampit=1}^{\nsamp} $ with early stopping.
    
    \vspace{0.2cm}
    \tcp{1.b. Training of input NN $h_{\parh}$: addressing surrogate problem}
    $\bullet$ Obtain the explicit embedding $\set{\embed_{\hat{\theta}}(y_{\sampit}) \in \sphere^{d-1}}_{\sampit=1}^{\nsamp}$ of the training output, and the ones $\set{\embed_{\hat{\theta}}(y_{\sampit}^{\mathrm{val}}) \in \sphere^{d-1}}_{\sampit=1}^{n_{\mathrm{val}}}$ of the validation output.

    $\bullet$ Solve $ \hat{\parh} = 
        \argmin_{\parh \in \Parh}\frac{1}{n} \sumsamp \norm{h_{\parh}(x_{\sampit}) - \embed_{\hat{\parembed}}(y_{\sampit})}{\surrogates}^2 $ by Adam optimizer, with early stopping based on mean squared error on validation set $\set{(x_{\sampit}^{\mathrm{val}}, \embed_{\hat{\theta}}(y_{\sampit}^{\mathrm{val}}))}_{\sampit=1}^{n_{\mathrm{val}}}$.
    
    \vspace{0.2cm}
    \tcp{2. Inference}
    $\bullet$ Obtain the explicit embedding $\set{\embed_{\hat{\parembed}}(y_i^{\mathrm{c}}) \in \sphere^{d-1}}_{\sampit=1}^{\nsamp_{\mathrm{c}}}$ of the candidates set

    $\bullet$ Initialize the predictions $\set{y^{(0)}_{\sampit}}_{i=1}^{n_{\mathrm{te}}}$ of the test inputs $\{x_i^{\mathrm{te}}\}_{i=1}^{n_{\mathrm{te}}}$ with the best candidate: 
    
    $\set{y^{(0)}_{\sampit} = y_{\hat{j}}^c \mid \hat{j} = \argmax_{1 \leq j \leq n_c }  \inner{\embed_{\hat{\theta}}(y_j^c)}{h_{\hat{\parh}}(x_i^{\mathrm{te}})}{\reals^d}}_{i=1}^{n_{\mathrm{te}}}$

    $\bullet$ Update the predictions by $T_2$ steps of projected gradient descent to solve the problem: $\forall \sampit \in \intset{1, n_{\mathrm{te}}}, \; \hat{y}_{\sampit} =  \mathcal{R}^{-1} \left(\argmin_{\yrelax\in \outputsrelax} \norm{\embed_{\hat{\parembed}}(\yrelax) - h_{\hat{\parh}}(x)}{\reals^d}^2 \right)$.
    \vspace{0.2cm}
\end{algorithm}
The framework \framework{} is summarized in Algorithm \ref{alg:didor}.

\section{Hyperparameters Details}
\label{app:hyper_para}

Table \ref{tab:hp_smi2mol_cl}, Table \ref{tab:hp_smi2mol_sr} and Table \ref{tab:hp_smi2mol_gd} present the details about the values of the hyperparameters or their search ranges on SMI2Mol dataset with \framework{} framework.

\begin{table*}[t!]
\caption{Contrastive learning hyper-parameters or their searching ranges during the cross-validation on SMI2Mol dataset.}
\begin{center}
\begin{tabular}{lc}
\toprule
Hyper-parameters &  Searching ranges  \\
\midrule
Node dropping rate &  $0.05$ \\
SGD batch size &  512 \\
SGD learning rate &  $10^{-3}$ \\
Number of R-GCN layers & $\set{2, 4, 6}$\\
Hidden dimension of R-GCN & $\set{32, 64, 128, 256, 512}$\\
Number of pre-training steps & 640,000\\

\bottomrule
\end{tabular}
\end{center}
\label{tab:hp_smi2mol_cl}
\end{table*}

\begin{table*}[t!]
\caption{Surrogate regression hyper-parameters or their searching ranges during the cross-validation on SMI2Mol dataset.}
\begin{center}
\begin{tabular}{lc}
\toprule
Hyper-parameters &  Searching ranges  \\
\midrule
SGD batch size &  128 \\
SGD learning rate &  $10^{-3}$ \\
Number of Transformer layers & $\set{2, 4, 6}$\\
Hidden dimension of Transformer& $\set{32, 64, 128, 256, 512}$\\
Number of heads of Attention in Transformer & $\set{2, 4, 8}$\\
Dropout rate of Transformer layers  & 0\\

\bottomrule
\end{tabular}
\end{center}
\label{tab:hp_smi2mol_sr}
\end{table*}

\begin{table*}[t!]
\caption{PGDBD hyper-parameters or their searching ranges during the cross-validation on SMI2Mol dataset.}
\begin{center}
\begin{tabular}{lc}
\toprule
Hyper-parameters &  Searching ranges  \\
\midrule
PGD number of steps &  \set{2000, 4000, 6000, 8000, 10000, 12000} \\
PGD step size &  $\set{0.1, 1, 10}$ \\

\bottomrule
\end{tabular}
\end{center}
\label{tab:hp_smi2mol_gd}
\end{table*}

%% file: arxiv.bbl
\begin{thebibliography}{60}
\providecommand{\natexlab}[1]{#1}
\providecommand{\url}[1]{\texttt{#1}}
\expandafter\ifx\csname urlstyle\endcsname\relax
  \providecommand{\doi}[1]{doi: #1}\else
  \providecommand{\doi}{doi: \begingroup \urlstyle{rm}\Url}\fi

\bibitem[Bakir et~al.(2007)Bakir, Hofmann, Schoelkopf, Smola, Taskar, and Vishwanathan]{bakir_predicting_2007}
G.~Bakir, T.~Hofmann, B.~Schoelkopf, A.~J. Smola, B.~Taskar, and S.~Vishwanathan, editors.
\newblock \emph{Predicting {Structured} {Data}}.
\newblock The MIT Press, July 2007.
\newblock ISBN 978-0-262-25569-1.

\bibitem[Belanger and McCallum(2016)]{belanger_structured_2016}
D.~Belanger and A.~McCallum.
\newblock Structured {Prediction} {Energy} {Networks}.
\newblock In M.~F. Balcan and K.~Q. Weinberger, editors, \emph{Proceedings of {The} 33rd {International} {Conference} on {Machine} {Learning}}, volume~48 of \emph{Proceedings of {Machine} {Learning} {Research}}, pages 983--992, New York, New York, USA, June 2016. PMLR.

\bibitem[Bellet et~al.(2013)Bellet, Habrard, and Sebban]{bellet_survey_2013}
A.~Bellet, A.~Habrard, and M.~Sebban.
\newblock A {Survey} on {Metric} {Learning} for {Feature} {Vectors} and {Structured} {Data}.
\newblock \emph{CoRR}, abs/1306.6709, 2013.
\newblock arXiv: 1306.6709.

\bibitem[Borgwardt et~al.(2020)Borgwardt, Ghisu, Llinares-López, O’Bray, and Rieck]{borgwardt_graph_2020}
K.~Borgwardt, E.~Ghisu, F.~Llinares-López, L.~O’Bray, and B.~Rieck.
\newblock Graph {Kernels}: {State}-of-the-{Art} and {Future} {Challenges}.
\newblock \emph{Foundations and Trends® in Machine Learning}, 13\penalty0 (5-6):\penalty0 531--712, 2020.
\newblock ISSN 1935-8237.

\bibitem[Brogat-Motte et~al.(2022)Brogat-Motte, Flamary, Brouard, Rousu, and D'Alché-Buc]{brogat-motte_learning_2022}
L.~Brogat-Motte, R.~Flamary, C.~Brouard, J.~Rousu, and F.~D'Alché-Buc.
\newblock Learning to {Predict} {Graphs} with {Fused} {Gromov}-{Wasserstein} {Barycenters}.
\newblock In K.~Chaudhuri, S.~Jegelka, L.~Song, C.~Szepesvari, G.~Niu, and S.~Sabato, editors, \emph{Proceedings of the 39th {International} {Conference} on {Machine} {Learning}}, volume 162 of \emph{Proceedings of {Machine} {Learning} {Research}}, pages 2321--2335. PMLR, July 2022.

\bibitem[Brouard et~al.(2016)Brouard, Szafranski, and d'Alché Buc]{brouard_input_2016}
C.~Brouard, M.~Szafranski, and F.~d'Alché Buc.
\newblock Input {Output} {Kernel} {Regression}: {Supervised} and {Semi}-{Supervised} {Structured} {Output} {Prediction} with {Operator}-{Valued} {Kernels}.
\newblock \emph{Journal of Machine Learning Research}, 17\penalty0 (176):\penalty0 1--48, 2016.

\bibitem[Chechik et~al.(2010)Chechik, Sharma, Shalit, and Bengio]{chechik_large_2010}
G.~Chechik, V.~Sharma, U.~Shalit, and S.~Bengio.
\newblock Large {Scale} {Online} {Learning} of {Image} {Similarity} {Through} {Ranking}.
\newblock \emph{J. Mach. Learn. Res.}, 11:\penalty0 1109--1135, Mar. 2010.
\newblock ISSN 1532-4435.
\newblock Publisher: JMLR.org.

\bibitem[Chen et~al.(2020)Chen, Kornblith, Norouzi, and Hinton]{chen_simple_2020}
T.~Chen, S.~Kornblith, M.~Norouzi, and G.~Hinton.
\newblock A {Simple} {Framework} for {Contrastive} {Learning} of {Visual} {Representations}.
\newblock In H.~D. III and A.~Singh, editors, \emph{Proceedings of the 37th {International} {Conference} on {Machine} {Learning}}, volume 119 of \emph{Proceedings of {Machine} {Learning} {Research}}, pages 1597--1607. PMLR, July 2020.

\bibitem[Chopra et~al.(2005)Chopra, Hadsell, and LeCun]{chopra_learning_2005}
S.~Chopra, R.~Hadsell, and Y.~LeCun.
\newblock Learning a {Similarity} {Metric} {Discriminatively}, with {Application} to {Face} {Verification}.
\newblock In \emph{2005 {IEEE} {Computer} {Society} {Conference} on {Computer} {Vision} and {Pattern} {Recognition} ({CVPR}'05)}, volume~1, pages 539--546 vol. 1, 2005.

\bibitem[Ciliberto et~al.(2016)Ciliberto, Rosasco, and Rudi]{ciliberto_consistent_2016}
C.~Ciliberto, L.~Rosasco, and A.~Rudi.
\newblock A {Consistent} {Regularization} {Approach} for {Structured} {Prediction}.
\newblock In D.~Lee, M.~Sugiyama, U.~Luxburg, I.~Guyon, and R.~Garnett, editors, \emph{Advances in {Neural} {Information} {Processing} {Systems}}, volume~29. Curran Associates, Inc., 2016.

\bibitem[Ciliberto et~al.(2020)Ciliberto, Rosasco, and Rudi]{ciliberto_general_2020}
C.~Ciliberto, L.~Rosasco, and A.~Rudi.
\newblock A {General} {Framework} for {Consistent} {Structured} {Prediction} with {Implicit} {Loss} {Embeddings}.
\newblock \emph{Journal of Machine Learning Research}, 21\penalty0 (98):\penalty0 1--67, 2020.

\bibitem[Cortes et~al.(2005)Cortes, Mohri, and Weston]{cortes_general_2005}
C.~Cortes, M.~Mohri, and J.~Weston.
\newblock A general regression technique for learning transductions.
\newblock In L.~D. Raedt and S.~Wrobel, editors, \emph{Machine {Learning}, {Proceedings} of the {Twenty}-{Second} {International} {Conference} ({ICML} 2005), {Bonn}, {Germany}, {August} 7-11, 2005}, volume 119 of \emph{{ACM} {International} {Conference} {Proceeding} {Series}}, pages 153--160. ACM, 2005.

\bibitem[Cortes et~al.(2010)Cortes, Mohri, and Rostamizadeh]{cortes_two-stage_2010}
C.~Cortes, M.~Mohri, and A.~Rostamizadeh.
\newblock Two-{Stage} {Learning} {Kernel} {Algorithms}.
\newblock In \emph{Proceedings of the 27th {Annual} {International} {Conference} on {Machine} {Learning} ({ICML} 2010)}, 2010.

\bibitem[Devlin et~al.(2019)Devlin, Chang, Lee, and Toutanova]{devlin_bert_2019}
J.~Devlin, M.-W. Chang, K.~Lee, and K.~Toutanova.
\newblock {BERT}: {Pre}-training of {Deep} {Bidirectional} {Transformers} for {Language} {Understanding}.
\newblock In \emph{Proceedings of the 2019 {Conference} of the {North} {American} {Chapter} of the {Association} for {Computational} {Linguistics}: {Human} {Language} {Technologies} ({NACCL}-{HLT})}, pages 4171--4186, June 2019.

\bibitem[Duchi et~al.(2008)Duchi, Shalev-Shwartz, Singer, and Chandra]{duchi_efficient_2008}
J.~Duchi, S.~Shalev-Shwartz, Y.~Singer, and T.~Chandra.
\newblock Efficient {Projections} onto the l1-{Ball} for {Learning} in {High} {Dimensions}.
\newblock In \emph{Proceedings of the 25th {International} {Conference} on {Machine} {Learning}}, {ICML} '08, pages 272--279, New York, NY, USA, 2008. Association for Computing Machinery.
\newblock ISBN 978-1-60558-205-4.
\newblock event-place: Helsinki, Finland.

\bibitem[El~Ahmad et~al.(2024{\natexlab{a}})El~Ahmad, Brogat-Motte, Laforgue, and d'Alché Buc]{el_ahmad_sketch_2024}
T.~El~Ahmad, L.~Brogat-Motte, P.~Laforgue, and F.~d'Alché Buc.
\newblock Sketch {In}, {Sketch} {Out}: {Accelerating} both {Learning} and {Inference} for {Structured} {Prediction} with {Kernels}.
\newblock In S.~Dasgupta, S.~Mandt, and Y.~Li, editors, \emph{Proceedings of {The} 27th {International} {Conference} on {Artificial} {Intelligence} and {Statistics}}, volume 238 of \emph{Proceedings of {Machine} {Learning} {Research}}, pages 109--117. PMLR, May 2024{\natexlab{a}}.

\bibitem[El~Ahmad et~al.(2024{\natexlab{b}})El~Ahmad, Yang, Laforgue, and d’Alché Buc]{el_ahmad_deep_2024}
T.~El~Ahmad, J.~Yang, P.~Laforgue, and F.~d’Alché Buc.
\newblock Deep {Sketched} {Output} {Kernel} {Regression} for {Structured} {Prediction}.
\newblock In A.~Bifet, J.~Davis, T.~Krilavičius, M.~Kull, E.~Ntoutsi, and I.~Žliobaitė, editors, \emph{Machine {Learning} and {Knowledge} {Discovery} in {Databases}. {Research} {Track}}, pages 93--110, Cham, 2024{\natexlab{b}}. Springer Nature Switzerland.
\newblock ISBN 978-3-031-70352-2.

\bibitem[Erhan et~al.(2010)Erhan, Bengio, Courville, Manzagol, Vincent, and Bengio]{erhan_why_2010}
D.~Erhan, Y.~Bengio, A.~Courville, P.-A. Manzagol, P.~Vincent, and S.~Bengio.
\newblock Why {Does} {Unsupervised} {Pre}-training {Help} {Deep} {Learning}?
\newblock \emph{Journal of Machine Learning Research}, 11\penalty0 (19):\penalty0 625--660, 2010.

\bibitem[Fink and Reymond(2007)]{fink_virtual_2007}
T.~Fink and J.-L. Reymond.
\newblock Virtual {Exploration} of the {Chemical} {Universe} up to 11 {Atoms} of {C}, {N}, {O}, {F}: {Assembly} of 26.4 {Million} {Structures} (110.9 {Million} {Stereoisomers}) and {Analysis} for {New} {Ring} {Systems}, {Stereochemistry}, {Physicochemical} {Properties}, {Compound} {Classes}, and {Drug} {Discovery}.
\newblock \emph{Journal of Chemical Information and Modeling}, 47\penalty0 (2):\penalty0 342--353, Mar. 2007.
\newblock ISSN 1549-9596.
\newblock Publisher: American Chemical Society.

\bibitem[Fink et~al.(2005)Fink, Bruggesser, and Reymond]{fink_virtual_2005}
T.~Fink, H.~Bruggesser, and J.-L. Reymond.
\newblock Virtual {Exploration} of the {Small}-{Molecule} {Chemical} {Universe} below 160 {Daltons}.
\newblock \emph{Angewandte Chemie International Edition}, 44\penalty0 (10):\penalty0 1504--1508, Feb. 2005.
\newblock ISSN 1433-7851.
\newblock Publisher: John Wiley \& Sons, Ltd.

\bibitem[Gao et~al.(2021)Gao, Yao, and Chen]{gao_simcse_2021}
T.~Gao, X.~Yao, and D.~Chen.
\newblock {SimCSE}: {Simple} {Contrastive} {Learning} of {Sentence} {Embeddings}.
\newblock In M.-F. Moens, X.~Huang, L.~Specia, and S.~W.-t. Yih, editors, \emph{Proceedings of the 2021 {Conference} on {Empirical} {Methods} in {Natural} {Language} {Processing}}, pages 6894--6910, Online and Punta Cana, Dominican Republic, Nov. 2021. Association for Computational Linguistics.

\bibitem[Gartner(2008)]{gaertner_2008}
T.~Gartner.
\newblock \emph{Kernels for structured data}, volume~72.
\newblock World Scientific, 2008.

\bibitem[Geurts et~al.(2006)Geurts, Wehenkel, and d'Alché Buc]{geurts_kernelizing_2006}
P.~Geurts, L.~Wehenkel, and F.~d'Alché Buc.
\newblock Kernelizing the {Output} of {Tree}-{Based} {Methods}.
\newblock In \emph{Proceedings of the 23rd {International} {Conference} on {Machine} {Learning}}, {ICML} '06, pages 345--352, New York, NY, USA, 2006. Association for Computing Machinery.
\newblock ISBN 1-59593-383-2.
\newblock event-place: Pittsburgh, Pennsylvania, USA.

\bibitem[Gilmer et~al.(2017)Gilmer, Schoenholz, Riley, Vinyals, and Dahl]{gilmer_neural_2017}
J.~Gilmer, S.~S. Schoenholz, P.~F. Riley, O.~Vinyals, and G.~E. Dahl.
\newblock Neural {Message} {Passing} for {Quantum} {Chemistry}.
\newblock In \emph{Proceedings of the 34th {International} {Conference} on {Machine} {Learning} - {Volume} 70}, {ICML}'17, pages 1263--1272. JMLR.org, 2017.
\newblock Place: Sydney, NSW, Australia.

\bibitem[Graber et~al.(2018)Graber, Meshi, and Schwing]{graber_deep_2018}
C.~Graber, O.~Meshi, and A.~Schwing.
\newblock Deep {Structured} {Prediction} with {Nonlinear} {Output} {Transformations}.
\newblock In S.~Bengio, H.~Wallach, H.~Larochelle, K.~Grauman, N.~Cesa-Bianchi, and R.~Garnett, editors, \emph{Advances in {Neural} {Information} {Processing} {Systems}}, volume~31. Curran Associates, Inc., 2018.

\bibitem[Grill et~al.(2020)Grill, Strub, Altché, Tallec, Richemond, Buchatskaya, Doersch, Avila~Pires, Guo, Gheshlaghi~Azar, Piot, kavukcuoglu, Munos, and Valko]{grill_bootstrap_2020}
J.-B. Grill, F.~Strub, F.~Altché, C.~Tallec, P.~Richemond, E.~Buchatskaya, C.~Doersch, B.~Avila~Pires, Z.~Guo, M.~Gheshlaghi~Azar, B.~Piot, k.~kavukcuoglu, R.~Munos, and M.~Valko.
\newblock Bootstrap {Your} {Own} {Latent} - {A} {New} {Approach} to {Self}-{Supervised} {Learning}.
\newblock In H.~Larochelle, M.~Ranzato, R.~Hadsell, M.~F. Balcan, and H.~Lin, editors, \emph{Advances in {Neural} {Information} {Processing} {Systems}}, volume~33, pages 21271--21284. Curran Associates, Inc., 2020.

\bibitem[Grover and Leskovec(2016)]{grover_node2vec_2016}
A.~Grover and J.~Leskovec.
\newblock node2vec: {Scalable} {Feature} {Learning} for {Networks}.
\newblock In \emph{Proceedings of the 22nd {ACM} {SIGKDD} {International} {Conference} on {Knowledge} {Discovery} and {Data} {Mining}}, {KDD} '16, pages 855--864, New York, NY, USA, 2016. Association for Computing Machinery.
\newblock ISBN 978-1-4503-4232-2.
\newblock event-place: San Francisco, California, USA.

\bibitem[Hagberg et~al.(2008)Hagberg, Schult, and Swart]{hagberg_exploring_2008}
A.~A. Hagberg, D.~A. Schult, and P.~J. Swart.
\newblock Exploring {Network} {Structure}, {Dynamics}, and {Function} using {NetworkX}.
\newblock In G.~Varoquaux, T.~Vaught, and J.~Millman, editors, \emph{Proceedings of the 7th {Python} in {Science} {Conference}}, pages 11 -- 15, Pasadena, CA USA, 2008.

\bibitem[Hassani and Khasahmadi(2020)]{hassani_contrastive_2020}
K.~Hassani and A.~H. Khasahmadi.
\newblock Contrastive {Multi}-{View} {Representation} {Learning} on {Graphs}.
\newblock In H.~D. III and A.~Singh, editors, \emph{Proceedings of the 37th {International} {Conference} on {Machine} {Learning}}, volume 119 of \emph{Proceedings of {Machine} {Learning} {Research}}, pages 4116--4126. PMLR, July 2020.

\bibitem[Ju et~al.(2024)Ju, Wang, Qin, Mao, Xiao, Luo, Yang, Gu, Wang, Long, Yi, Luo, and Zhang]{ju_survey_2024}
W.~Ju, Y.~Wang, Y.~Qin, Z.~Mao, Z.~Xiao, J.~Luo, J.~Yang, Y.~Gu, D.~Wang, Q.~Long, S.~Yi, X.~Luo, and M.~Zhang.
\newblock Towards graph contrastive learning: A survey and beyond, 2024.

\bibitem[Kadri et~al.(2013)Kadri, Ghavamzadeh, and Preux]{kadri_generalized_2013}
H.~Kadri, M.~Ghavamzadeh, and P.~Preux.
\newblock A {Generalized} {Kernel} {Approach} to {Structured} {Output} {Learning}.
\newblock In S.~Dasgupta and D.~McAllester, editors, \emph{Proceedings of the 30th {International} {Conference} on {Machine} {Learning}}, volume~28 of \emph{Proceedings of {Machine} {Learning} {Research}}, pages 471--479, Atlanta, Georgia, USA, June 2013. PMLR.
\newblock Issue: 1.

\bibitem[Kingma and Ba(2015)]{kingma_adam_2015}
D.~P. Kingma and J.~Ba.
\newblock Adam: {A} {Method} for {Stochastic} {Optimization}.
\newblock In \emph{International {Conference} on {Learning} {Representations} ({ICLR})}, May 2015.

\bibitem[Kipf and Welling(2017)]{kipf_semi-supervised_2017}
T.~N. Kipf and M.~Welling.
\newblock Semi-{Supervised} {Classification} with {Graph} {Convolutional} {Networks}.
\newblock In \emph{International {Conference} on {Learning} {Representations} ({ICLR})}, 2017.

\bibitem[Lafferty et~al.(2001)Lafferty, McCallum, and Pereira]{lafferty_conditional_2001}
J.~D. Lafferty, A.~McCallum, and F.~C.~N. Pereira.
\newblock Conditional {Random} {Fields}: {Probabilistic} {Models} for {Segmenting} and {Labeling} {Sequence} {Data}.
\newblock In \emph{Proceedings of the {Eighteenth} {International} {Conference} on {Machine} {Learning}}, {ICML} '01, pages 282--289, San Francisco, CA, USA, 2001. Morgan Kaufmann Publishers Inc.
\newblock ISBN 1-55860-778-1.

\bibitem[Le-Khac et~al.(2020)Le-Khac, Healy, and Smeaton]{le-khac_contrastive_2020}
P.~H. Le-Khac, G.~Healy, and A.~F. Smeaton.
\newblock Contrastive {Representation} {Learning}: {A} {Framework} and {Review}.
\newblock \emph{IEEE Access}, 8:\penalty0 193907--193934, 2020.

\bibitem[Liu et~al.(2019)Liu, Ott, Goyal, Du, Joshi, Chen, Levy, Lewis, Zettlemoyer, and Stoyanov]{liu_roberta:_2019}
Y.~Liu, M.~Ott, N.~Goyal, J.~Du, M.~Joshi, D.~Chen, O.~Levy, M.~Lewis, L.~Zettlemoyer, and V.~Stoyanov.
\newblock {RoBERTa}: {A} {Robustly} {Optimized} {BERT} {Pretraining} {Approach}.
\newblock \emph{arXiv preprint arXiv:1907.11692}, July 2019.
\newblock arXiv: 1907.11692.

\bibitem[Logeswaran and Lee(2018)]{logeswaran_efficient_2018}
L.~Logeswaran and H.~Lee.
\newblock An {Efficient} {Framework} for {Learning} {Sentence} {Representations}.
\newblock In \emph{International {Conference} on {Learning} {Representations}}, 2018.

\bibitem[Luise et~al.(2019)Luise, Stamos, Pontil, and Ciliberto]{luise_leveraging_2019}
G.~Luise, D.~Stamos, M.~Pontil, and C.~Ciliberto.
\newblock Leveraging {Low}-{Rank} {Relations} {Between} {Surrogate} {Tasks} in {Structured} {Prediction}.
\newblock In K.~Chaudhuri and R.~Salakhutdinov, editors, \emph{Proceedings of the 36th {International} {Conference} on {Machine} {Learning}}, volume~97 of \emph{Proceedings of {Machine} {Learning} {Research}}, pages 4193--4202. PMLR, June 2019.

\bibitem[Mahoney(2011)]{mahoney_randomized_2011}
M.~W. Mahoney.
\newblock Randomized {Algorithms} for {Matrices} and {Data}.
\newblock \emph{Found. Trends Mach. Learn.}, 3\penalty0 (2):\penalty0 123--224, Feb. 2011.
\newblock ISSN 1935-8237.
\newblock Place: Hanover, MA, USA Publisher: Now Publishers Inc.

\bibitem[Micchelli and Pontil(2005)]{micchelli_learning_2005}
C.~A. Micchelli and M.~Pontil.
\newblock On {Learning} {Vector}-{Valued} {Functions}.
\newblock \emph{Neural Computation}, 17\penalty0 (1):\penalty0 177--204, Jan. 2005.
\newblock ISSN 0899-7667.
\newblock \_eprint: https://direct.mit.edu/neco/article-pdf/17/1/177/816069/0899766052530802.pdf.

\bibitem[Mnih and Kavukcuoglu(2013)]{mnih_learning_2013}
A.~Mnih and K.~Kavukcuoglu.
\newblock Learning {Word} {Embeddings} {Efficiently} with {Noise}-{Contrastive} {Estimation}.
\newblock In C.~J. Burges, L.~Bottou, M.~Welling, Z.~Ghahramani, and K.~Q. Weinberger, editors, \emph{Advances in {Neural} {Information} {Processing} {Systems}}, volume~26. Curran Associates, Inc., 2013.

\bibitem[Nowak et~al.(2019)Nowak, Bach, and Rudi]{nowak_sharp_2019}
A.~Nowak, F.~Bach, and A.~Rudi.
\newblock Sharp {Analysis} of {Learning} with {Discrete} {Losses}.
\newblock In K.~Chaudhuri and M.~Sugiyama, editors, \emph{Proceedings of the {Twenty}-{Second} {International} {Conference} on {Artificial} {Intelligence} and {Statistics}}, volume~89 of \emph{Proceedings of {Machine} {Learning} {Research}}, pages 1920--1929. PMLR, Apr. 2019.

\bibitem[Nowozin and Lampert(2011)]{nowozin_structured_2011}
S.~Nowozin and C.~H. Lampert.
\newblock \emph{Structured learning and prediction in computer vision}, volume~6.
\newblock Now publishers Inc, 2011.

\bibitem[Oord et~al.(2018)Oord, Li, and Vinyals]{oord_representation_2018}
A.~v.~d. Oord, Y.~Li, and O.~Vinyals.
\newblock Representation {Learning} with {Contrastive} {Predictive} {Coding}.
\newblock \emph{arXiv preprint arXiv:1807.03748}, 2018.

\bibitem[Radford et~al.(2018)Radford, Narasimhan, Salimans, and Sutskever]{radford_improving_2018}
A.~Radford, K.~Narasimhan, T.~Salimans, and I.~Sutskever.
\newblock Improving {Language} {Understanding} by {Generative} {Pre}-{Training}.
\newblock Technical report, OpenAI, 2018.
\newblock URL \url{https://openai.com/blog/language-unsupervised/}.

\bibitem[Ramakrishnan et~al.(2014)Ramakrishnan, Dral, Rupp, and von Lilienfeld]{ramakrishnan_quantum_2014}
R.~Ramakrishnan, P.~O. Dral, M.~Rupp, and O.~A. von Lilienfeld.
\newblock Quantum chemistry structures and properties of 134 kilo molecules.
\newblock \emph{Scientific Data}, 1, 2014.
\newblock Publisher: Nature Publishing Group.

\bibitem[Reimers and Gurevych(2019)]{reimers_sentence-bert_2019}
N.~Reimers and I.~Gurevych.
\newblock Sentence-{BERT}: {Sentence} {Embeddings} using {Siamese} {BERT}-{Networks}.
\newblock In \emph{Proceedings of the 2019 {Conference} on {Empirical} {Methods} in {Natural} {Language} {Processing} and the 9th {International} {Joint} {Conference} on {Natural} {Language} {Processing} ({EMNLP}-{IJCNLP})}, pages 3980--3990, Hong Kong, China, 2019. Association for Computational Linguistics.

\bibitem[Ruddigkeit et~al.(2012)Ruddigkeit, van Deursen, Blum, and Reymond]{ruddigkeit_enumeration_2012}
L.~Ruddigkeit, R.~van Deursen, L.~C. Blum, and J.-L. Reymond.
\newblock Enumeration of 166 {Billion} {Organic} {Small} {Molecules} in the {Chemical} {Universe} {Database} {GDB}-17.
\newblock \emph{Journal of Chemical Information and Modeling}, 52\penalty0 (11):\penalty0 2864--2875, Nov. 2012.
\newblock ISSN 1549-9596.
\newblock Publisher: American Chemical Society.

\bibitem[Schlichtkrull et~al.(2018)Schlichtkrull, Kipf, Bloem, van den Berg, Titov, and Welling]{schlichtkrull_modeling_2018}
M.~Schlichtkrull, T.~N. Kipf, P.~Bloem, R.~van den Berg, I.~Titov, and M.~Welling.
\newblock Modeling {Relational} {Data} with {Graph} {Convolutional} {Networks}.
\newblock In A.~Gangemi, R.~Navigli, M.-E. Vidal, P.~Hitzler, R.~Troncy, L.~Hollink, A.~Tordai, and M.~Alam, editors, \emph{The {Semantic} {Web}}, pages 593--607, Cham, 2018. Springer International Publishing.
\newblock ISBN 978-3-319-93417-4.

\bibitem[Schölkopf et~al.(1997)Schölkopf, Smola, and Müller]{scholkopf_kernel_1997}
B.~Schölkopf, A.~Smola, and K.-R. Müller.
\newblock Kernel {Principal} {Component} {Analysis}.
\newblock In W.~Gerstner, A.~Germond, M.~Hasler, and J.-D. Nicoud, editors, \emph{Artificial {Neural} {Networks} — {ICANN}'97}, pages 583--588, Berlin, Heidelberg, 1997. Springer Berlin Heidelberg.
\newblock ISBN 978-3-540-69620-9.

\bibitem[Sohn(2016)]{sohn_improved_2016}
K.~Sohn.
\newblock Improved {Deep} {Metric} {Learning} with {Multi}-class {N}-pair {Loss} {Objective}.
\newblock In D.~Lee, M.~Sugiyama, U.~Luxburg, I.~Guyon, and R.~Garnett, editors, \emph{Advances in {Neural} {Information} {Processing} {Systems}}, volume~29. Curran Associates, Inc., 2016.

\bibitem[Sun et~al.(2020)Sun, Hoffman, Verma, and Tang]{sun_infograph_2020}
F.-Y. Sun, J.~Hoffman, V.~Verma, and J.~Tang.
\newblock {InfoGraph}: {Unsupervised} and {Semi}-supervised {Graph}-{Level} {Representation} {Learning} via {Mutual} {Information} {Maximization}.
\newblock In \emph{International {Conference} on {Learning} {Representations}}, 2020.

\bibitem[Tsochantaridis et~al.(2005)Tsochantaridis, Joachims, Hofmann, and Altun]{tsochantaridis_large_2005}
I.~Tsochantaridis, T.~Joachims, T.~Hofmann, and Y.~Altun.
\newblock Large {Margin} {Methods} for {Structured} and {Interdependent} {Output} {Variables}.
\newblock \emph{Journal of Machine Learning Research}, 6\penalty0 (50):\penalty0 1453--1484, 2005.

\bibitem[Vaswani et~al.(2017)Vaswani, Shazeer, Parmar, Uszkoreit, Jones, Gomez, Kaiser, and Polosukhin]{vaswani_attention_2017}
A.~Vaswani, N.~Shazeer, N.~Parmar, J.~Uszkoreit, L.~Jones, A.~N. Gomez, L.~Kaiser, and I.~Polosukhin.
\newblock Attention is {All} you {Need}.
\newblock In \emph{Advances in {Neural} {Information} {Processing} {Systems} ({NIPS})}, pages 5998--6008, 2017.

\bibitem[Vayer et~al.(2019)Vayer, Courty, Tavenard, Laetitia, and Flamary]{vayer_optimal_2019}
T.~Vayer, N.~Courty, R.~Tavenard, C.~Laetitia, and R.~Flamary.
\newblock Optimal {Transport} for structured data with application on graphs.
\newblock In K.~Chaudhuri and R.~Salakhutdinov, editors, \emph{Proceedings of the 36th {International} {Conference} on {Machine} {Learning}}, volume~97 of \emph{Proceedings of {Machine} {Learning} {Research}}, pages 6275--6284. PMLR, June 2019.

\bibitem[Veličković et~al.(2019)Veličković, Fedus, Hamilton, Liò, Bengio, and Hjelm]{velickovic_deep_2019}
P.~Veličković, W.~Fedus, W.~L. Hamilton, P.~Liò, Y.~Bengio, and R.~D. Hjelm.
\newblock Deep {Graph} {Infomax}.
\newblock In \emph{International {Conference} on {Learning} {Representations}}, 2019.

\bibitem[Weston et~al.(2003)Weston, Chapelle, Vapnik, Elisseeff, and Schölkopf]{weston_kernel_2003}
J.~Weston, O.~Chapelle, V.~Vapnik, A.~Elisseeff, and B.~Schölkopf.
\newblock Kernel {Dependency} {Estimation}.
\newblock In S.~Becker, S.~Thrun, and K.~Obermayer, editors, \emph{Advances in {Neural} {Information} {Processing} {Systems}}, volume~15. MIT Press, 2003.

\bibitem[Woodruff(2014)]{woodruff_sketching_2014}
D.~P. Woodruff.
\newblock Sketching as a {Tool} for {Numerical} {Linear} {Algebra}.
\newblock \emph{Found. Trends Theor. Comput. Sci.}, 10\penalty0 (1–2):\penalty0 1--157, Oct. 2014.
\newblock ISSN 1551-305X.
\newblock Place: Hanover, MA, USA Publisher: Now Publishers Inc.

\bibitem[Yang et~al.(2024)Yang, Labeau, and d'Alché Buc]{yang_exploiting_2024}
J.~Yang, M.~Labeau, and F.~d'Alché Buc.
\newblock Exploiting {Edge} {Features} in {Graph}-based {Learning} with {Fused} {Network} {Gromov}-{Wasserstein} {Distance}.
\newblock \emph{Transactions on Machine Learning Research}, 2024.
\newblock ISSN 2835-8856.

\bibitem[You et~al.(2020)You, Chen, Sui, Chen, Wang, and Shen]{you_graph_2020}
Y.~You, T.~Chen, Y.~Sui, T.~Chen, Z.~Wang, and Y.~Shen.
\newblock Graph {Contrastive} {Learning} with {Augmentations}.
\newblock In H.~Larochelle, M.~Ranzato, R.~Hadsell, M.~F. Balcan, and H.~Lin, editors, \emph{Advances in {Neural} {Information} {Processing} {Systems}}, volume~33, pages 5812--5823. Curran Associates, Inc., 2020.

\end{thebibliography}
